\pdfoutput=1

\documentclass[11pt]{article}

\usepackage{acl}

\usepackage{times}
\usepackage{latexsym}
\usepackage{tabularx}
\usepackage{graphicx}
\usepackage{enumitem}
\usepackage{placeins}

\usepackage[T1]{fontenc}

\usepackage[utf8]{inputenc}

\usepackage{microtype}
\usepackage{booktabs}

\newcommand\vD{\mathcal{D}}
\newcommand\vU{\mathcal{U}}
\newcommand\vN{\mathcal{N}}
\newcommand\vC{\mathcal{C}}

\newcommand\vP{P}

\newcommand\vn{n}
\newcommand\vc{c}

\newcommand\base{\textsc{Base}}
\newcommand\ours{\textsc{User}}
\newcommand\gpt{\textsc{Gpt}}
\newcommand\cycle{\textsc{Cyc}}
\newcommand\grammar{\textsc{Gram}}
\newcommand\rerank{\textsc{Rank}}
\newcommand\iter{\textsc{3x}}

\title{Addressing Resource and Privacy Constraints in Semantic Parsing Through Data Augmentation}

\author{{\bf Kevin Yang}$^{1,*}$\ \ \ \ 
    {\bf Olivia Deng}$^2$\ \ \ \ 
  {\bf Charles Chen}$^2$\\
  {\bf Richard Shin}$^2$\ \ \ \ 
  {\bf Subhro Roy}$^2$\ \ \ \ 
  {\bf Benjamin Van Durme}$^2$\\
    $^1$UC Berkeley, $^2$Microsoft Semantic Machines \\
     \texttt{yangk@berkeley.edu},  \texttt{sminfo@microsoft.com}
     }

\begin{document}
\maketitle
\begin{abstract}


We introduce a novel setup for low-resource task-oriented semantic parsing which incorporates several constraints that may arise in real-world scenarios: (1) lack of similar datasets/models from a related domain, (2) inability to sample useful logical forms directly from a grammar, and (3) privacy requirements for unlabeled natural utterances. Our goal is to improve a low-resource semantic parser using utterances collected through user interactions. In this highly challenging but realistic setting, we investigate data augmentation approaches involving generating a set of structured canonical utterances corresponding to logical forms, before simulating corresponding natural language and filtering the resulting pairs. We find that such approaches are effective despite our restrictive setup: in a low-resource setting on the complex SMCalFlow calendaring dataset~\cite{andreas2020task}, we observe $33\%$ relative improvement over a non-data-augmented baseline in top-1 match.\footnotetext[1]{*: Work done during internship at Semantic Machines.}

\end{abstract}

\setlist[enumerate]{topsep=0pt,itemsep=-1ex,partopsep=1ex,parsep=1ex}

\section{Introduction}

We aim to improve the performance of a semantic parser based on previous user interactions, but without making use of their direct utterances, nor any associated personal identifiable information (PII). Such privacy requirements are common in practical deployment~\cite{kannan2016smart}, and semantic parsers are commonly used in real-world systems such as Siri and Alexa, converting natural language into structured queries to be executed downstream~\cite{kamath2018survey}. 


Constructing semantic parsers can be expensive: annotating examples consisting of natural language-logical form pairs often requires trained experts. 
Two complementary lines of work has pursued this concern. First, several works~\cite{zhong2020grounded,cao2020unsupervised} tackle \textit{low-resource} semantic parsing via approaches such as data augmentation.
A second line of work~\cite{wang2015building,xiao2016sequence} explores \textit{canonical utterances:} structured language which maps directly to logical forms, but resembles natural language (Table \ref{tab:smcalflow_example}). The use of canonical forms as the target of  semantic parsing has shown benefits in accuracy~\cite{shin2021constrained,wu2021paraphrasing}.

\begin{table}[!tbp]
\small
\centering
\begin{tabularx}{0.47\textwidth}{lX}
\toprule
\textbf{Natural} & {\it When is Allison's birthday?}\\
\midrule
\textbf{Logical} & (Yield :output (:start (singleton (:results (FindEventWrapperWithDefaults :constraint (Constraint[Event] :subject (?~= \#(String ``Allison's birthday'')))))))) \\
\midrule
\textbf{Canonical} & start time of find event called something like ``Allison's birthday''\\
\bottomrule
\end{tabularx}
\vspace{-2mm}
\caption{An example of natural language, logical form, and canonical form in the SMCalFlow domain. The event title, {\it ``Allison's birthday,''} is private information.
}
\label{tab:smcalflow_example}
\vspace{-1.5em}
\end{table}


We consider low-resource semantic parsing with further resource and privacy constraints which may arise in practical deployment: beyond a small gold dataset of labeled pairs, we assume only unlabeled natural utterances which must be masked for PII. Unlike many prior works, we assume that (1) we do not have a large dataset of related logical forms in a different domain, (2) we cannot sample arbitrarily many useful logical forms, and (3) we must the preserve privacy of user utterances. 



We propose several approaches which are compatible with our imposed restrictions, broadly following three steps: (1) generate a set of privacy-preserving canonical utterances; (2) simulate corresponding natural utterances; and (3) filter the resulting canonical-natural utterance pairs to yield additional ``silver'' data for training. We more than double the performance of a non-data-augmented baseline on the ATIS domain~\cite{hemphill1990atis}, and achieve a $33\%$ relative improvement on the more realistic SMCalFlow domain~\cite{andreas2020task}.
We hope these experiments help motivate further research interest in parser improvement for realistic scenarios.

\section{Semantic Parsing in Practice}


Our setup assumes access exclusively to:

\begin{enumerate}
    \item a small ``seed'' dataset $\vD$ of natural utterance with corresponding parses, and
    \item a larger set $\vU$ of unlabeled natural utterances, for which PII must be masked before use.
\end{enumerate}

In a real-world setting, one might hand-annotate the seed dataset $\vD$ to train a system for initial deployment, while then leveraging\, $\vU$ to refine a future version of the system. 


While our setting is highly restrictive,
we argue that it reflects practical constraints. For example, in practice, the grammar for logical forms---as well as the synchronous context-free grammar (SCFG) that maps them to canonical utterances---will often be written from scratch, precluding transfer learning methods which leverage a large quantity of similar data in another domain. 
Moreover, in complex domains, one cannot expect to \textit{sample} useful logical forms directly from a grammar if the grammar is designed for \textit{coverage} as in e.g., SMCalFlow~\cite{andreas2020task}. Therefore, other than $\vD$, the only additional data (excluding additional manual annotation) are subsequent user inputs in the form of $\vU$, with PII masked to preserve privacy.


\section{Related Work}








Compared to prior work in low-resource semantic parsing, our task setup's constraints require different approaches.

First, we consider semantic parsing on an entirely \textit{new} grammar for logical forms, rather than adapting to new domains starting from a \textit{preexisting} grammar~\cite{zhao2019data,zhong2020grounded,burnyshev2021single,kim2021neuralwoz,tseng2021transferable}. For example, \citet{zhong2020grounded} takes a natural-language-to-SQL model for one database to propose language-SQL training examples for another database.

Second, we assume one cannot sample useful canonical utterances directly from the grammar, unlike \citet{zhong2020grounded} and \citet{cao2020unsupervised}. For example, \citet{cao2020unsupervised} use a backtranslation-esque approach leveraging large numbers of unlabeled natural and canonical utterances. 

Moreover, we do not even assume direct access to unlabeled natural utterances, due to real-world privacy considerations~\cite{kannan2016smart,campagna2017almond}. Many works on low-resource semantic parsing, such as those mentioned previously, do not consider the privacy aspect.

Nevertheless, recent work~\cite{shin2021constrained,wu2021paraphrasing,yin2021ingredients,schucher2021power} has demonstrated decent performance given just a small seed dataset $\vD$, by combining pretrained language models with constrained decoding. For example, \citet{shin2021constrained} use only 300 labeled examples in the complex SMCalFlow dialogue domain~\cite{andreas2020task}. However, using pretrained models to directly \textit{generate} silver training data, with a method such as DINO~\cite{schick2021generating}, is unsuitable in semantic parsing: the models are unaware of either the underlying grammar or the space of parsable queries. 
One of our contributions is to explore more effective uses of pretrained models
for data augmentation in a practical semantic parsing scenario. 

Finally, the detection of PII in user data is an applied topic of interest~\cite{pilan2022text}, such as for safely summarizing call transcripts~\cite{amazon}
or the automatic detection of doxing~\cite{karimi2022automated}. In our work we implement a solution meant as a proof of concept for our exploration, based on detecting and replacing named entities. 




\section{Practical Augmentation}\label{sec:method}
\begin{figure}[t!]
    \centering
    \includegraphics[width=0.47\textwidth]{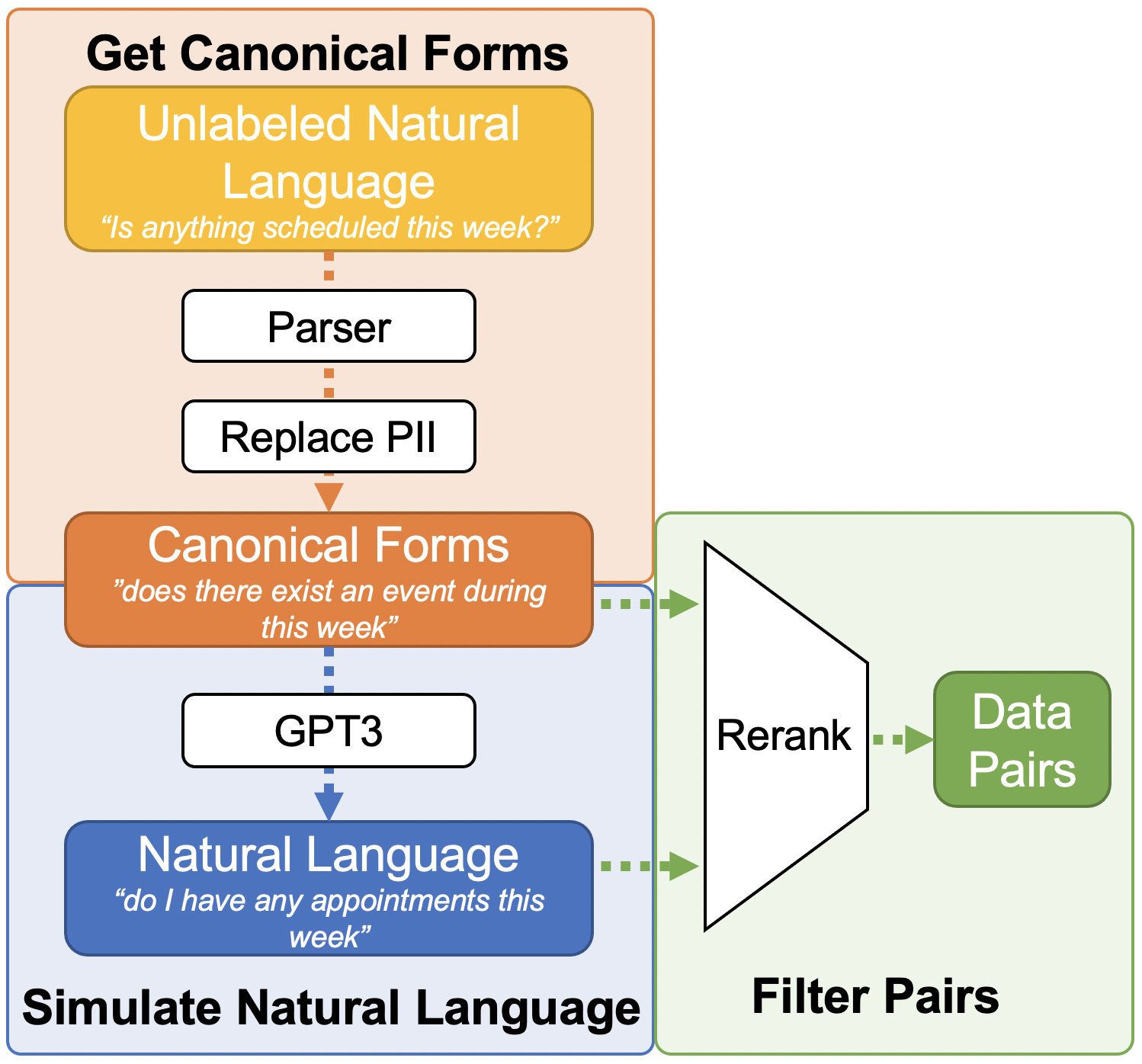}
    \caption{Illustration of one of our proposed methods for data augmentation (\ours{}-\rerank{}) in low-resource semantic parsing. We first obtain canonical forms from unlabeled user data using a parser trained on seed data, replacing PII. Next, we simulate corresponding natural language for the generated canonical forms. Finally, we filter the canonical-natural pairs to obtain our final silver data pairs for augmentation. 
    }
    \label{fig:method}
\end{figure}

While finetuning a pretrained model on the seed dataset $\vD$ can yield a reasonable parser $\vP$~\cite{shin2021constrained,wu2021paraphrasing}, we aim to increase performance via data augmentation. However, our realistic setup precludes many prior approaches. We propose to generate silver data via three main steps, shown in Figure \ref{fig:method}: (1) generate a set $\vC$ of canonical utterances $\vc$, (2) simulate a set $\vN$ of corresponding natural utterances $\vn$, and (3) filter the resulting $(\vc, \vn)$ pairs. We suggest multiple approaches for these steps, and benchmark their efficacy in Sec. \ref{sec:experiments}. The entire procedure can be iterated multiple times as the parser improves.

\subsection{Generating Canonical Utterances}\label{sec:generate_canonical}
First, we generate canonical utterances $\vc$. In principle, one could sample directly from a task-specific grammar, but the results may not be useful in practice (Sec. \ref{sec:experiments}). The remaining options are to generate $\vc$ conditioned on either unlabeled natural utterances $\vU$ or the seed data $\vD$. 

\medskip
\noindent\textbf{Generation conditioned on $\vU$} (\ours{}).
We need to mask all PII, but this is difficult to guarantee in the original natural language domain. Therefore, we first train a parser $\vP$ on $\vD$, and parse each utterance in $\vU$ to obtain a set of canonical utterances $\vC'$. In the more structured domain of $\vC'$ we can guarantee masking and replacing all PII to yield the final set $\vC$. Critically, it is not necessary that the initial $\vC'$ are correct parses of $\vU$; we only need a realistic distribution over canonical utterances, and the initial $\vU$ is no longer parallel to the final $\vC$ anyway due to replacing PII. Hence it is acceptable if the parser $\vP$'s errors are numerous but unbiased. In any case, the final $\vC$ will be somewhat tied to the true distribution of user utterances in $\vU$.


\medskip
\noindent\textbf{Generation conditioned on $\vD$} (\gpt{}). A second method of generating $\vC$ is SCFG-constrained decoding on an autoregressive language model,\footnote{Ideally we would use GPT3~\cite{brown2020language}, and we do so in the ATIS domain, but API limitations in GPT3 together with the requirements of our constrained decoding force us to use GPT2-XL~\cite{radford2019language} in SMCalFlow.} prompting with the seed data $\vD$. 
Specifically, we prompt with a random concatenation of plans from $\vD$, separated by newlines. The SCFG that defines canonical utterances constrains the decoding, forcing the model to output a valid canonical utterance. 


\subsection{Simulated Natural Utterances}

For each canonical $\vc$ in $\vC$, we now re-generate a natural utterance $\vn$. While other methods (e.g., finetuning) are possible, here we employ a prompting approach using GPT3~\cite{brown2020language}. We use a prompt containing $\vD$'s canonical-natural pairs, ending with the canonical utterance $\vc$ for which we want to sample a corresponding $\vn$. 

\subsection{Filtering Silver Data}\label{sec:filter}
Many $(\vc, \vn)$ pairs we generate may be low-quality, depending on the task and seed data $\vD$ available. To obtain more high-quality pairs, we simulate $20$ natural utterances $\vn$ for each $\vc$. We must then filter the resulting pairs, which we do based on either reranking or cycle consistency. 

\medskip
\noindent\textbf{Reranking} (\rerank{}). We accept the best of $20$ simulated $\vn$ for each $\vc$, and add this $(\vc, \vn)$ to our training data. The reranker combines two scores: (1) the log-probability that the original $\vD$-trained parser $\vP$ parses $\vn$ back to the original canonical $\vc$, and (2) the edit distance between $\vn$ and $\vc$ (capped based on the length of $\vc$), which should intuitively be \textit{maximized} to encourage linguistic diversity in the augmented data, perhaps at a small accuracy cost. 

\medskip
\noindent\textbf{Cycle consistency} (\cycle{}). We accept a $(\vc, \vn)$ pair if the original parser $\vP$ parses $\vn$ back to $\vc$. This assures the resulting pairs' quality, but may skew the distribution toward easier examples, which are less helpful in downstream training.

\section{Experiments}
\label{sec:experiments}

\textbf{Tasks.} We evaluate on two domains, both English:

\begin{enumerate}
    \item ATIS~\cite{hemphill1990atis}, a flight booking dataset. We use the Break~\cite{wolfson2020break} subset.\footnote{We also ran preliminary experiments on the DROP~\cite{dua2019drop} and NLVR2~\cite{suhr2018corpus} subsets of Break, but found that the canonical utterances were too unnatural for any method to perform reasonably (Appendix \ref{sec:other_datasets}).} 
    \item SMCalFlow~\cite{andreas2020task}, a calendaring dataset, which we view as the most complex and realistic.
\end{enumerate} 


In each domain, we assume a seed data $\vD$ of just 30 pairs, conducting several trials with different random samples of seed data to mitigate noise from this selection. We sample 300 unlabeled natural utterances $\vU$ from the dataset, which must be parsed to canonical forms (using the grammar and SCFG of \citet{shin2021constrained}) and then PII-masked before use. Our implementation of PII masking is based on recognizing and replacing named entities; see Appendix \ref{sec:masking_pii} for further details. 

\medskip
\noindent\textbf{Methods.} We evaluate several methods on each task, listed below. 

\begin{enumerate}
    \item \textit{\base{}}, a supervised baseline which finetunes BART~\cite{lewis2019bart} on the seed $\vD$ following \citet{shin2021constrained}, discarding $\vU$.
    \item \textit{\ours{}-\rerank{}}, a data augmentation approach following the \ours{} and \rerank{} methods described in Sec. \ref{sec:generate_canonical} and \ref{sec:filter} respectively, and depicted in Figure \ref{fig:method}. 
    \item \textit{\gpt{}-\rerank{}}, a similar approach which generates $\vc$ following \gpt{} from Sec. \ref{sec:generate_canonical} instead. 
    \item \textit{\ours{}-\cycle{}}, a version which filters $(\vc, \vn)$ pairs via cycle consistency (Sec. \ref{sec:filter}).
    \item \textit{\grammar{}-\rerank{}}, a weak baseline that samples initial $\vc$ directly from the grammar, which we run only on SMCalFlow since our ATIS grammar is too loosely specified for sampling.
\end{enumerate}


\noindent\textbf{Results.} We observe that our best data augmentation methods (\ours{}-\rerank{}, \gpt{}-\rerank{}) double the performance of the baseline finetuning method \base{} on ATIS, and outperform it on SMCalFlow by up to $20\%$ relative gain (Table \ref{tab:smcalflow}).\footnote{Although the standard deviations appear large, the variation between trials is largely due to randomness in selecting the seed data $\vD$. For example, \ours{}-\rerank{} is better than \base{} on SMCalFlow with $p=.0004$ on a paired $t$-test.} Nonetheless, absolute performance remains low due to the tiny amount of seed data, although we note that the exact match metric may be unnecessarily harsh, penalizing some semantically equivalent parses. 

Of interest is that \gpt{}-\rerank{} outperforms \base{} despite using only the seed $\vD$, and not extra unlabeled $\vU$. Moreover, iterating the data augmentation procedure (\ours{}-\rerank{}-\iter{}, \gpt{}-\rerank{}-\iter{}) can further improve performance compared to \base{} (relative 150\% on ATIS, 33\% relative on SMCalFlow), by improving the initial parser $\vP$ used for parsing unlabeled $\vU$ or for filtering pairs $(\vc, \vn)$, although we observed in preliminary experiments that further iterations yielded diminishing benefits. 

In contrast, \ours{}-\cycle{} performs poorly on ATIS, indicating that the \cycle{} filtering is perhaps too restrictive for certain domains. Even on SMCalFlow where performance is decent in comparison, the successful cycles are overwhelmingly for relatively trivial canonical utterances (e.g., ``Hello! How are you?''). We additionally observe that nearly one-third of cycles are successful, much more than the actual validation set accuracy of 15\%, indicating that our auto-generated user utterances remain less challenging and diverse compared to real user utterances. Meanwhile, \grammar{}-\rerank{} is no better than the unaugmented baseline \base{}: sampling plans directly from a grammar is ineffective in a complex, realistic domain like SMCalFlow.

\begin{table}[]
\centering
\
\begin{tabular}{lcc}
\toprule
\textbf{Method}                     & \textbf{ATIS} & \textbf{SMCalFlow}  \\
\midrule
\base{}   & \phantom{0}6.8 $\pm$ 3.5 & 13.2 $\pm$ 3.4 \\
\ours{}-\rerank{}  & 13.4 $\pm$ 4.1 & 15.5 $\pm$ 3.7 \\
\gpt{}-\rerank{}   & 13.7 $\pm$ 3.2 & 15.9 $\pm$ 2.7 \\
\ours{}-\cycle{}  & \phantom{0}6.0 $\pm$ 2.3 & 15.0 $\pm$ 4.0\\
\grammar{}-\rerank{} &  & 13.4 $\pm$ 2.8 \\
\midrule
\ours{}-\rerank{}-\iter{} & \textbf{17.3 $\pm$ 1.3} & \textbf{17.6 $\pm$ 4.6} \\
\gpt{}-\rerank{}-\iter{}  & 16.7 $\pm$ 3.5 & 16.1 $\pm$ 3.0 \\
\bottomrule
\end{tabular}
\caption{\textit{Main results on ATIS and SMCalFlow for different methods.} Top-1 parsing match percentage evaluated over 5 (ATIS) or 10 (SMCalFlow) trials on different seed datasets $\vD$. For the two highest-performing methods, \ours{}-\rerank{} and \gpt{}-\rerank{}, we iterate data augmentation 3 times on SMCalFlow, yielding \ours{}-\rerank{}-\iter{} and \gpt{}-\rerank{}-\iter{}. \ours{}-\rerank{}-\iter{} performs best overall.}
\label{tab:smcalflow}
\end{table}



\subsection{Analysis}

We conduct additional analyses on SMCalFlow. 

\medskip
\noindent\textbf{Reranking.} First, we run ablations on reranking in \ours{}-\rerank{} (Table \ref{tab:reranking}). While our edit distance heuristic described in Sec. \ref{sec:filter} makes little difference,
reranking of some form is crucial. Meanwhile, there are many possibilities for other reranking procedures.

\begin{table}[htbp]
\centering
\begin{tabular}{lc}
\toprule
\textbf{Method}                      & \textbf{SMCalFlow}  \\
\midrule
\base{}    & 13.2 $\pm$ 3.4\\
\ours{}-\rerank{}-\iter{}  & 17.6 $\pm$ 4.6 \\
\ours{}-\textsc{NoEditRank}-\iter{} & 17.3 $\pm$ 4.7 \\
\ours{}-\textsc{NoRank} & 12.8 $\pm$ 3.5 \\
\bottomrule
\end{tabular}
\caption{\textit{SMCalFlow reranking ablations.} Since the version without reranking (\ours{}-\textsc{NoRank}) is no better than the baseline, we do not iterate the data augmentation procedure. The edit distance heuristic makes little difference in this case (\ours{}-\textsc{NoEditRank}-\iter{} vs. \ours{}-\rerank{}-\iter{}), but reranking is crucial. }
\label{tab:reranking}
\end{table}

\medskip
\noindent\textbf{Effect of Masking PII}. We rerun our full pipeline for \ours{}-\rerank{} on SMCalFlow, removing only the step where we resampled PII, in order to isolate the effect of doing so (Table \ref{tab:pii_ablation}, \ours{}-\rerank{}-\textsc{KeepPII}). As one might expect, replacing PII hurts performance, albeit slightly. Of course, if PII is not a concern, then many other data augmentation schemes from prior work become possible again.


\begin{table}[htbp]
\centering

\begin{tabular}{lc}
\toprule
\textbf{Method}                      & \textbf{SMCalFlow}  \\
\midrule
\ours{}-\rerank{}  & 15.5 $\pm$ 3.7 \\
\ours{}-\rerank{}-\textsc{KeepPII}  & 16.2 $\pm$ 2.8\\
\bottomrule
\end{tabular}
\caption{\textit{SMCalFlow ablation where we do not resample PII.} As expected, performance is slightly better if we do not need to resample PII.}
\label{tab:pii_ablation}
\end{table}

\medskip
\noindent\textbf{Additional Seed Data.} We explore using a larger seed dataset $\vD$ on both ATIS and SMCalFlow (90 and 100 data points respectively, instead of 30). On SMCalFlow, we observe that \ours{}-\rerank{}'s gains over the baseline largely disappear (Table \ref{tab:100}). Thus, improved data augmentation methods which still yield gains with larger seed datasets are an important direction for future exploration.

\begin{table}[htbp]
\centering
\small
\begin{tabular}{@{}lcc@{}}
\toprule
\textbf{Method}            &    \textbf{ATIS (90 seed)}      & \textbf{SMCalFlow (100 seed)}  \\
\midrule
\base{}  & 21.4 $\pm$ 1.8 & 31.6 $\pm$ 0.3\\
\ours{}-\rerank{} & 21.4 $\pm$ 1.7 & 31.7 $\pm$ 1.0 \\
\bottomrule
\end{tabular}
\caption{\textit{Results with more seed data.} We use a seed dataset $\vD$ of size 90 (ATIS) or 100 (SMCalFlow) rather than 30, with 3 trials per method. The gains from data augmentation largely disappear at this scale, so we do not do additional augmentation iterations.}
\label{tab:100}
\end{table}



\medskip
\noindent\textbf{Examples and Error Analysis.} Finally, in Table \ref{tab:main_examples} we show several SMCalFlow example parses by the baseline \base{} compared to our highest-performing method \ours{}-\rerank{}-\iter{}. Compared to \base{}, \ours{}-\rerank{}-\iter{} is often better at segmenting names (Example 1), and is also more likely to be semantically similar to the gold parse in cases where \base{} is wildly incorrect (Example 2). Nevertheless, in the latter example, \ours{}-\rerank{}-\iter{} is still marked wrong, suggesting that our exact match metric may somewhat underrepresent the performance of all models. Finally, both methods struggle on more complex and/or composite intents (Example 3). Additional examples illustrating these phenomena are shown in Appendix \ref{sec:examples}.

\newpage

\begin{table}[h]
\small
\centering
\begin{tabularx}{0.47\textwidth}{lX}
\toprule
\textbf{Example 1}\\
\cmidrule(lr){1-2}
Previous Agent & \\
\cmidrule(lr){1-2}
Natural & {\it please make a meet with my doctor sarah}\\
\cmidrule(lr){1-2}
\base{} & create event with " doctor"\\
\cmidrule(lr){1-2}
\ours{}-\rerank{}-\iter{} & create event with " doctor sarah"\\
\cmidrule(lr){1-2}
Gold Canonical & create event with " doctor sarah"\\
\toprule
\textbf{Example 2}\\
\cmidrule(lr){1-2}
Previous Agent & {\it Let me know if there's anything else I can help you with.}\\
\cmidrule(lr){1-2}
Natural & {\it no}\\
\cmidrule(lr){1-2}
\base{} & does there exist an event tomorrow 9 military\\
\cmidrule(lr){1-2}
\ours{}-\rerank{}-\iter{} & Looks good!\\
\cmidrule(lr){1-2}
Gold Canonical & Thanks for your help!\\
\toprule
\textbf{Example 3}\\
\cmidrule(lr){1-2}
Previous Agent & \\
\cmidrule(lr){1-2}
Natural & {\it Please accept the bowling fundraiser and tell sammy I will bring refreshments.}\\
\cmidrule(lr){1-2}
\base{} & create event called " bowling fundraiser" starting month 4 11 2019 5 PM\\
\cmidrule(lr){1-2}
\ours{}-\rerank{}-\iter{} & create event called " bowling fundraiser"\\
\cmidrule(lr){1-2}
Gold Canonical & respond Accepted with comment " I will bring refreshments" to find event called something like " bowling fundraiser"\\
\bottomrule
\end{tabularx}
\vspace{-2mm}
\caption{Example parses by the baseline \base{} and our best method \ours{}-\rerank{}-\iter{} on SMCalFlow. Each example contains the previous agent utterance (if it exists) and user utterance in the first two lines, followed by the \base{} parse, \ours{}-\rerank{}-\iter{} parse, and gold parse.
}
\vspace{-1.5em}
\label{tab:main_examples}
\end{table}

\section{Conclusion}

We have discussed a challenging setting for low resource semantic parsing based on real-world resource and privacy constraints. In addition to a seed dataset, the only resources allowed are unlabeled natural utterances which must be PII-masked. We observe that data augmentation approaches leveraging pretrained language models can still improve over supervised baselines which use only the seed dataset. At the same time, substantial room remains for  improvement: there are many alternatives to our reranking procedure for silver data, and our method loses some effectiveness when more labeled data is provided. We hope that our exploratory observations help lay a foundation for further work in realistic data augmentation approaches for semantic parsing. 



\clearpage

\section*{Ethical Considerations}

We believe our work makes a positive impact by focusing heavily on the need for privacy considerations when exploring low-resource settings for semantic parsing. However, as our methods rely heavily on large pretrained language models such as GPT3, we may inherit similar biases which such models are known for~\cite{brown2020language}. 
\section*{Acknowledgements}

We thank the rest of the team at Semantic Machines, as well as our anonymous reviewers, for their support and helpful comments which aided us greatly in improving the paper. We also thank the NSF for their support through a fellowship to the first author.

\bibliography{anthology,custom}
\bibliographystyle{acl_natbib}

\clearpage

\appendix



\section{Masking and Replacing Personal Identifiable Information}
\label{sec:masking_pii}

\subsection{ATIS}

The ATIS grammar is somewhat loosely defined and does not clearly indicate the instances of PII. This would be problematic in a real production setting due to making it difficult to guarantee masking out all PII. However, for our experiments we simply truecase the data and apply named entity recognition using spaCy~\cite{Honnibal_spaCy_Industrial-strength_Natural_2020}, which we find is highly successful from a qualitative inspection. We treat detected named entities as PII. 

\begin{figure}[htbp]
    \centering
    \includegraphics[width=0.5\textwidth]{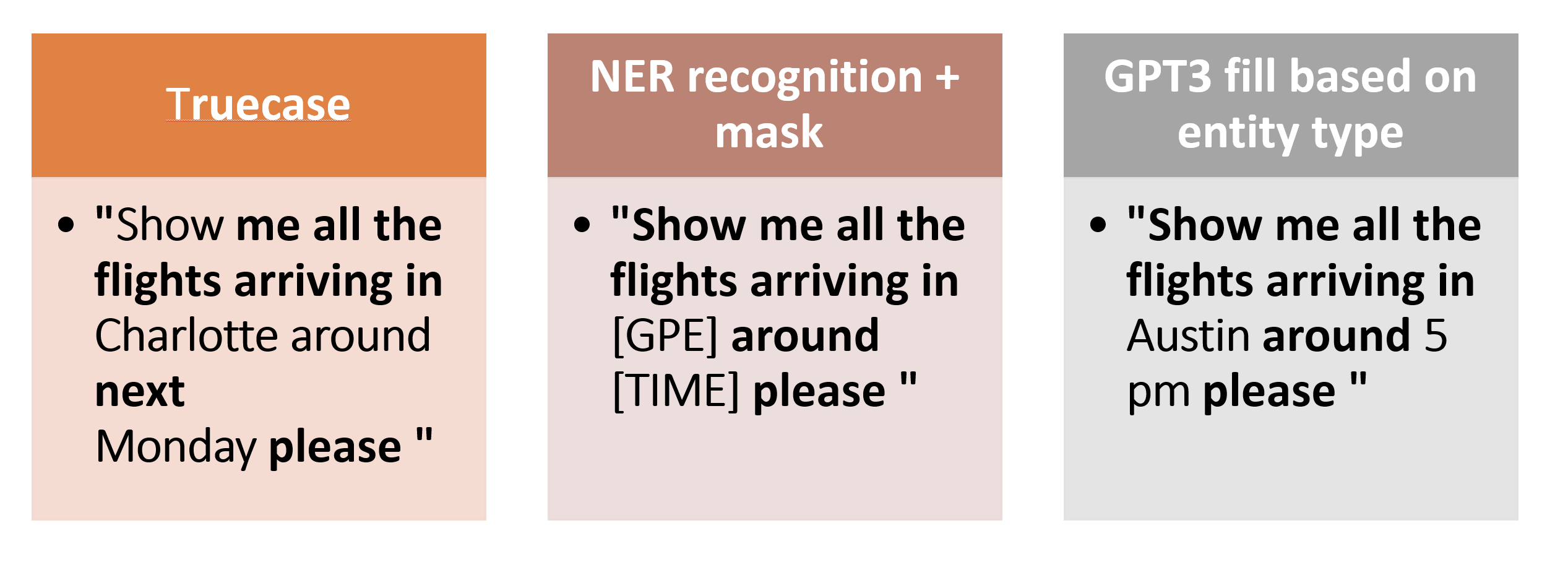}
    \caption{An example of infill masking PII.
    }
    \label{fig:infill_mask_pii}
\end{figure}

To remove PII, we devise two methods: 1. masking and 2. generating entirely new plans. In 1., we apply the above method to detect PII, mask it with the entity type, and ask GPT3 to infill (Figure \ref{fig:infill_mask_pii}). In 2., correponding to our \gpt{}-\rerank{} method, we feed GPT3 example plans from the seed data and ask for an entirely new plan that does not contain PII (Figure \ref{fig:gpt_plans}). 

\begin{figure}[htbp]
    \centering

\begin{dialogue}
\speak{input} 

return ground transportation ;return \#1 in boston ;return \#2 between the airport ;return \#3 and  downtown \\\vspace{1em}
return flights ;return \#1 that  are nonstop ;return \#2 from san diego ;return \#3 to  new york \\\vspace{1em}

return flights ;return \#1 from  denver ;return \#2 to  pittsburgh ;return \#3 leaving after 6pm ;return \#3 leaving before 7pm ;return \#1 of  both \#4 and \#5 \\\vspace{1em}

\speak{gpt3} \textbf{return flights ;return \#1 that  are nonstop ;return \#2 between denver ;return \#3 and  oakland ;return \#4 leaving after noon ;return \#5 arriving after 5pm}

\end{dialogue}
    \caption{INPUT is an example prompt, consisting of several plans from the seed data, given to GPT3 to generate new plans to begin the data augmentation procedure in \gpt{}-\rerank{}. 
    }
    \label{fig:gpt_plans}
\end{figure}

\subsection{SMCalFlow}

Since the SMCalFlow grammar (described in detail in Appendix A.2 of \citet{shin2021constrained}) is type-annotated, we define three categories of PII: names, event titles, and locations. Each category is easily identifiable from the logical form, so it suffices to sample a new value from the same category in the logical form to guarantee that PII is replaced. 

We sample names from a distribution balanced for ethnicity and gender. For event titles and locations, we sample them from GPT3 by prompting with seed data canonical forms containing event titles and/or locations, and then prefixing the generation with \texttt{find event called something like "} (event titles) or a mix of \texttt{weather at "} and \texttt{find event at "} (locations). We cut off the generation once the next \texttt{"} appears.

\section{Example Parses}

\label{sec:examples}

We include some additional example parses for \ours{}-\rerank{}-\iter{} compared to \base{} on SMCalFlow, as well as \gpt{}-\rerank{}-\iter{} compared to \base{} on ATIS, when both start with the same seed data (Tables \ref{tab:smcalflow_comparison_examples}, \ref{tab:smcalflow_comparison_examples2}, \ref{tab:atis_comparison_examples}). 

We additionally provide some qualitative analysis on the SMCalFlow examples. Most examples that both methods get correct are relatively simple intents (Example 1). However, there are also many examples where one or both methods output a parse which a human might judge semantically equivalent to the true parse (Examples 2 and 3), suggesting that our exact match metric might underrepresent the true strength of the models. 

The next few examples illustrate cases in which \ours{}-\rerank{}-\iter{} improves over \base{}, for example by being better at segmenting names (Examples 4). \base{} is also sometimes just wildly incorrect on examples where \ours{}-\rerank{}-\iter{} matches the gold exactly, or at least closely in meaning (Examples 5 and 6). Of course, given the tiny amount of initial training data, \ours{}-\rerank{}-\iter{} is not immune to these types of errors either (Example 7).

\begin{table}[!tbp]
\small
\centering
\begin{tabularx}{0.47\textwidth}{lX}
\toprule
\textbf{Example 1}\\
\cmidrule(lr){1-2}
Previous Agent & {\it I've deleted your event matching "doctor's appointment".}\\
\cmidrule(lr){1-2}
Natural & {\it thank you}\\
\cmidrule(lr){1-2}
\base{} & Thanks for your help!\\
\cmidrule(lr){1-2}
\ours{}-\rerank{}-\iter{} & Thanks for your help!\\
\cmidrule(lr){1-2}
Gold Canonical & Thanks for your help!\\
\toprule
\textbf{Example 2}\\
\cmidrule(lr){1-2}
Previous Agent & {\it Ok, I won't do anything for now. Let me know what else I can help you with.}\\
\cmidrule(lr){1-2}
Natural & {\it Make an appointment for tomorrow from 1pm to 6pm}\\
\cmidrule(lr){1-2}
\base{} & create event starting tomorrow 1 PM to 6 PM\\
\cmidrule(lr){1-2}
\ours{}-\rerank{}-\iter{} & create event starting tomorrow 1 PM to 6 PM\\
\cmidrule(lr){1-2}
Gold Canonical & create event starting tomorrow 1 PM ending 6 PM after that datetime\\
\toprule
\textbf{Example 3}\\
\cmidrule(lr){1-2}
Previous Agent & {\it Next is your tour potential development sites on Thursday the 28th from 3:00 to 5:00 PM.}\\
\cmidrule(lr){1-2}
Natural & {\it I want to delete that one.}\\
\cmidrule(lr){1-2}
\base{} & delete the event\\
\cmidrule(lr){1-2}
\ours{}-\rerank{}-\iter{} & delete find event\\
\cmidrule(lr){1-2}
Gold Canonical & delete the event\\
\toprule
\textbf{Example 4}\\
\cmidrule(lr){1-2}
Previous Agent & {\it Does one of these work?}\\
\cmidrule(lr){1-2}
Natural & {\it When is Easter next year?}\\
\cmidrule(lr){1-2}
\base{} & start time of find event called something like " Easter next year"\\
\cmidrule(lr){1-2}
\ours{}-\rerank{}-\iter{} & start time of find event called something like " Easter" starting next year\\
\cmidrule(lr){1-2}
Gold Canonical & Easter next year\\

\bottomrule
\end{tabularx}
\vspace{-2mm}
\caption{Example parses by the baseline \base{} and our best method \ours{}-\rerank{}-\iter{} on SMCalFlow.
}
\label{tab:smcalflow_comparison_examples}
\vspace{-1.5em}
\end{table}

\begin{table}[htbp]
\small
\centering
\begin{tabularx}{0.47\textwidth}{lX}

\toprule
\textbf{Example 5}\\
\cmidrule(lr){1-2}
Previous Agent & \\
\cmidrule(lr){1-2}
Natural & {\it list to me my calendar please}\\
\cmidrule(lr){1-2}
\base{} & create event on today afternoon\\
\cmidrule(lr){1-2}
\ours{}-\rerank{}-\iter{} & find event\\
\cmidrule(lr){1-2}
Gold Canonical & find event\\
\toprule
\textbf{Example 6}\\
\cmidrule(lr){1-2}
Previous Agent & {\it The "library" is on Monday the 30th from 10:00 to 10:30 AM.}\\
\cmidrule(lr){1-2}
Natural & {\it Ok! now tell me when does my Coffee Date start?}\\
\cmidrule(lr){1-2}
\base{} & ERROR: can't answer trivia\\
\cmidrule(lr){1-2}
\ours{}-\rerank{}-\iter{} & start time of find event called something like " Coffee Date"\\
\cmidrule(lr){1-2}
Gold Canonical & start time of find event called something like " Coffee Date"\\
\toprule
\textbf{Example 7}\\
\cmidrule(lr){1-2}
Previous Agent & \\
\cmidrule(lr){1-2}
Natural & {\it  Hey, I was wondering who the organizer is for the museum event next week.}\\
\cmidrule(lr){1-2}
\base{} & ERROR: can't answer trivia\\
\cmidrule(lr){1-2}
\ours{}-\rerank{}-\iter{} & ERROR: can't answer trivia\\
\cmidrule(lr){1-2}
Gold Canonical & organizer of find event called something like " museum" during next week\\
\bottomrule
\end{tabularx}
\vspace{-2mm}
\caption{Additional example parses by the baseline \base{} and our best method \ours{}-\rerank{}-\iter{} on SMCalFlow.
}
\label{tab:smcalflow_comparison_examples2}
\vspace{-3em}
\end{table}

\begin{table}[htbp]
\small
\centering
\begin{tabularx}{0.47\textwidth}{lX}
\toprule
\textbf{Example 8}\\
\cmidrule(lr){1-2}
Previous Agent & \\
\cmidrule(lr){1-2}
Natural & {\it I want a flight from houston to memphis on tuesday morning}\\
\cmidrule(lr){1-2}
\base{} & return flights ;return \#1 from houston ;return \#2 to memphis ;return \#3 on tuesday morning\\
\cmidrule(lr){1-2}
\gpt{}-\rerank{}-\iter{} & return flights ;return \#1 from houston ;return \#2 to memphis ;return \#3 on tuesday morning\\
\cmidrule(lr){1-2}
Gold Canonical & return flights ;return \#1 from houston ;return \#2 to memphis ;return \#3 on tuesday ;return \#4 in the morning\\
\toprule
\textbf{Example 9}\\
\cmidrule(lr){1-2}
Previous Agent & \\
\cmidrule(lr){1-2}
Natural & {\it What ground transportation is available from the pittsburgh airport to downtown and how much does it cost}\\
\cmidrule(lr){1-2}
\base{} & return transportation ;return \#1 that is ground ;return \#2 to downtown pittsurgh; return cost of \#4 \\
\cmidrule(lr){1-2}
\gpt{}-\rerank{}-\iter{} & return transportation ;return \#1 that is ground ;return \#2 from the pittsburgh ;return \#3 to downtown pittsburgh; return cost \#4 \\
\cmidrule(lr){1-2}
Gold Canonical & return ground transportation ;return \#1 which  is  available ;return \#2 from  the pittsburgh airport ;return \#3 to downtown ;return the cost of \#4\\

\bottomrule
\end{tabularx}
\vspace{-2mm}
\caption{Additional example parses by the baseline \base{} and our best method \gpt{}{}-\rerank{}-\iter{} on ATIS.
}
\label{tab:atis_comparison_examples}
\vspace{-1.5em}
\end{table}

\section{Preliminary Experiments on Other Break Subsets}
\label{sec:other_datasets}

We additionally ran preliminary experiments on the DROP~\cite{dua2019drop} (reading comprehension) and NLVR2~\cite{suhr2018corpus} (language-vision reasoning) subsets of Break~\cite{wolfson2020break}. We used a similar setup to our ATIS and SMCalFlow experiments, with 30 initial seed data $\vD$ and 300 unlabeled user utterances $\vU$. 

However, across multiple trials of multiple methods (\base{}, \ours{}-\rerank{}, \gpt{}-\rerank{}, \ours{}-\cycle{}), we never observed performance above 2\% on either domain. This may be partially due to the diversity of the data; for example, DROP is an amalgamation of data from several sources. However, we hypothesize that this across-the-board poor performance is primarily the result of an SCFG for canonical utterances which results in somewhat unnatural language (Table \ref{tab:drop_nlvr2_example}), and that performance could be greatly improved with a better SCFG. Given the current form of our canonical utterances in DROP and NLVR2, it is challenging to learn the task given just 30 seed examples. In comparison, the SMCalFlow canonical utterances (Table \ref {tab:smcalflow_example} in the main text) are much more natural. 

\begin{table}[!htbp]
\small
\centering
\begin{tabularx}{0.47\textwidth}{lX}
\toprule
\textbf{DROP Natural} & \textit{Which player had the shortest touchdown reception of the game?}\\
\cmidrule(lr){1-2}
\textbf{DROP Canonical} & return touchdown receptions ;return shortest of \#1 ;return player of \#2 \\
\toprule
\textbf{NLVR2 Natural} & \textit{If there are two carts, but only one of them has a canopy.}\\
\cmidrule(lr){1-2}
\textbf{NLVR2 Canonical} & return carts ;return number of  \#1 ;return if  \#2 is equal to  two ;return canopy ;return \#1 that has \#4 ;return number of  \#5 ;return if  \#6 is equal to  one ;return if  both  \#3 and \#7 are true \\
\bottomrule
\end{tabularx}
\caption{Examples of natural utterances with corresponding canonical utterances for DROP and NLVR2 domains. The language of the canonical utterances is relatively unnatural.}
\label{tab:drop_nlvr2_example}
\end{table}

We additionally inspect some inaccurate example predictions by \base{} on DROP and NLVR2, which are often wildly incorrect (Table \ref{tab:drop_nlvr2_pred}). We also show some example $(\vc, \vn)$ pairs generated by our data augmentation procedure, demonstrating the failure to propose good natural language $\vn$ given the limited data and unnatural canonical $\vc$ (Table \ref{tab:drop_nlvr2_su}). 

\begin{table}[!htbp]
\small
\centering
\begin{tabularx}{0.47\textwidth}{lX}
\toprule
\textbf{DROP Natural} & \textit{Which player threw more yards in the game, Young or Manning?} \\
\cmidrule(lr){1-2}
\textbf{DROP Top-1 Parse} & return that was the highest ;return that was more of \#1 ;return number of \#2 for each \#1 ;return \#1 where \#3 is lower than one ;return number of \#4 \\
\toprule
\textbf{NLVR2 Natural} & \textit{If there are bananas with stickers on them}\\
\cmidrule(lr){1-2}
\textbf{NLVR2 Top-1 Parse} 
&return, ;return number of \#1 ;return if \#2 is equal to one \\
\bottomrule
\end{tabularx}
\caption{Predictions by \base{} on DROP and NLVR2 which are wildly incorrect. Our data augmentation methods fare no better.}
\label{tab:drop_nlvr2_pred}
\end{table}

\begin{table}[!htbp]
\small
\centering
\begin{tabularx}{0.47\textwidth}{lX}
\toprule
\textbf{DROP Canonical} & return the five \\
\cmidrule(lr){1-2}
\textbf{DROP Simulated Natural} & Fact-checkers failed to catch five factual errors. \\
\toprule
\textbf{NLVR2 Canonical} & return left image ;return \#1 that are dirty ;return if \#2 is in one of the images\\
\cmidrule(lr){1-2}
\textbf{NLVR2 Simulated Natural} & If any of the trucks are dirty.\\
\bottomrule
\end{tabularx}
\caption{Example simulated natural utterances generated by prompting GPT3 on DROP and NLVR2, after reranking and selecting the best of 20 generations. The correspondence between canonical and simulated natural utterances remains imperfect.}
\label{tab:drop_nlvr2_su}
\end{table}

\end{document}